\documentclass[11pt]{article}

\usepackage[a4paper,margin=1in]{geometry}
\usepackage[utf8]{inputenc}
\usepackage[T1]{fontenc}
\usepackage{lmodern}
\usepackage{microtype}
\usepackage{parskip}
\usepackage{amsmath,amssymb,amsthm,bm}
\usepackage{mathtools}

\usepackage{graphicx}
\usepackage{booktabs}
\usepackage{multirow}
\usepackage{array}
\usepackage{caption}
\usepackage{subcaption}
\usepackage{float}

\usepackage{xcolor}
\definecolor{linkblue}{RGB}{15,80,160}
\usepackage[colorlinks=true,linkcolor=linkblue,citecolor=linkblue,urlcolor=linkblue]{hyperref}
\usepackage{url}

\usepackage[numbers,sort&compress]{natbib}

\usepackage{listings}
\title{%
    \textbf{Shape: A Self-Supervised 3D Geometry Foundation Model for Industrial CAD Analysis}
}

\author{%
    Bayangmbe Mounmo$^{1}$ \quad
    Sam Chien$^{1}$ \quad
    Mile Mitrovic$^{2}$ \\[0.5em]
    $^{1}$SIMD AI \quad\quad $^{2}$SB AI Lab
} 

\date{}

\begin{document}
\maketitle

\begin{abstract}
Industrial CAD workflows require robust, generalizable 3D geometric representations supporting accuracy and explainability. We introduce \texttt{Shape}, a self-supervised foundation model converting surface meshes into dense per-token embeddings. Shape combines a structured 3D latent grid, a multi-scale geometry-aware tokenizer (MAGNO) with cross-attention, and a transformer processor using grouped-query attention and RMSNorm. A learned reconstruction prior enables per-region attribution for explainable predictions. Pretraining uses masked-token reconstruction of normalized geometry statistics and multi-resolution contrastive consistency. The 10.9M-parameter backbone is pretrained on 61,052 CAD meshes from Thingi10K, MFCAD, and Fusion360. On a held-out split of 2{,}983 meshes, \texttt{Shape} achieves reconstruction $R^2 = 0.729$ and $98.1\%$ top-1 retrieval under the Wang–Isola protocol, with near-zero reconstruction train/val gap (contrastive scores use a larger evaluation pool). A $2 \times 2$ ablation on loss type and target-space normalization shows per-dimension normalization is critical: without it, performance collapses ($R^2 < 0.14$, top-1 $< 88\%$); with it, both losses succeed ($R^2 > 0.70$, top-1 $> 96\%$). Smooth-L1 offers secondary stability. Code, embeddings, and an interactive demo are released at \url{https://github.com/simd-ai/shape}.
\end{abstract}

\section{Introduction}
\label{sec:introduction}

Industrial computational fluid dynamics (CFD) and finite element analysis (FEA) workflows rely heavily on exploiting geometric structure to keep simulations tractable. Mirror symmetry lets an analyst simulate half a jet engine combustor; axisymmetry reduces a turbine stage to a 2D mesh; cyclic sectors collapse a bladed disk to a single blade passage. Missing these reductions can inflate the compute cost of a single CFD run by an order of magnitude; finding them by hand requires domain expertise and remains one of the dominant time sinks in simulation setup.

Despite clear demand, to our knowledge there is no openly released self-supervised foundation model for industrial 3D CAD geometry that (i) learns transferable per-token representations from raw meshes without relying on expensive expert labels, (ii) provides a principled attribution mechanism that maps model outputs back to specific regions of the input mesh surface, and (iii) is small and fast enough to serve as the starting point for domain-specific downstream tasks such as symmetry detection, primitive recognition, and simulation-reduction recommendation. The closest prior work -- Point-BERT~\cite{yu2022pointbert} and Point-MAE~\cite{pang2022pointmae} -- target generic object point clouds, not industrial CAD meshes with the heavy-tailed curvature distributions that characterize machined parts.

\paragraph{Contributions.} The main contributions of this
work are articulated as follows:

\begin{enumerate}
\item \textbf{Architecture.} We adopt the \emph{Geometry Aware Operator Transformer} (GAOT) of Wen et al.~\cite{wen2025gaot}, originally proposed for PDE surrogate modeling on arbitrary domains, and repurpose it as the backbone of a self-supervised foundation model for 3D CAD geometry. The backbone composes (a) a Multiscale Attentional Graph Neural Operator (MAGNO)~\cite{wen2025gaot} tokenizer that cross-attends from a structured 3D latent grid to physical surface points using cosine-similarity attention with learned temperature and multi-scale neighborhood statistics, and (b) a transformer processor~\cite{vaswani2017attention} with grouped-query attention~\cite{ainslie2023gqa}, RMSNorm~\cite{zhang2019rmsnorm}, and optional rotary position embeddings~\cite{su2021rope}. The architecture decouples mesh resolution from token count: any mesh, regardless of its vertex count, is encoded into the same fixed-size token grid. To our knowledge, this is the first self-supervised application of GAOT/MAGNO; the original work targets supervised PDE surrogate training.

\item \textbf{Training recipe and controlled ablation.} We identify \emph{calibrated per-dimension target normalization} as the single decisive intervention for stable pretraining on heavy-tailed CAD statistics. We compute per-dimension mean and standard deviation of the reconstruction targets once on the training split, store them as non-trainable buffers, and apply them as a z-score before the regression loss. Without this normalization, the masked-token reconstruction loss is approximately four orders of magnitude larger than the other pretraining terms and training is dominated by a handful of curvature dimensions regardless of whether the loss is MSE or Smooth-L1. A $2 \times 2$ controlled ablation (Section~\ref{subsec:ablation}) isolates this effect: both MSE and Smooth-L1 fail with $R^2 < 0.14$ on raw targets, and both succeed with $R^2 > 0.70$ with normalization. The choice of Smooth-L1 over MSE is a secondary stability decision we retain for robustness over longer training horizons, but it is not the cause of the large performance swing.

\item \textbf{Empirical validation.} We pretrain a 10.9\,M-parameter model on 61{,}052 CAD meshes from three industrial-focused datasets (Thingi10K, MFCAD, Fusion360) using 8\,$\times$\,H100 80\,GB GPUs for 50 epochs. On a held-out hash-based validation split of 2{,}983 meshes, we report a reconstruction $R^2 = 0.729$, a contrastive top-1 retrieval accuracy of $98.1\%$ under the Wang--Isola~\cite{wang2020understanding} protocol, and effectively zero train/val gap on the masked-token reconstruction objective.

\item \textbf{Built-in explainability.} We show that the learned reconstruction projection head naturally produces a per-token attribution map: after pretraining, the squared error between predicted and normalized target at each token, rendered back onto the mesh surface, highlights geometrically novel or out-of-distribution regions. This attribution mechanism does not require any post-hoc method such as LRP~\cite{bach2015lrp}; it falls out of the pretraining objective directly.
\end{enumerate}

We release the code, pretrained weights, precomputed embeddings, and an interactive 3D demo of masked-token reconstruction and shape retrieval under a permissive license. A companion scaling roadmap targets 300\,M and 1\,B parameter versions over the next two training generations.

\section{Related Work}
\label{sec:related}

\paragraph{Foundation models and self-supervised pretraining.}
The transfer of transformer-based self-supervised learning from language~\cite{devlin2019bert,radford2018gpt} to vision~\cite{dosovitskiy2020vit,he2022mae} established a blueprint of large unlabeled corpora paired with a denoising or reconstruction objective. Masked Autoencoders (MAE)~\cite{he2022mae} showed that masking 75\% of image patches and reconstructing their pixels yields representations that transfer to a wide range of downstream tasks; similar masking ratios have been explored in 3D. DINOv2~\cite{oquab2024dinov2} combines multiple self-supervised signals (discriminative and reconstructive) at scale to produce all-purpose visual features. Our work applies the same philosophy to industrial CAD meshes: a structured latent grid, a masked-token objective on geometric statistics (rather than coordinates), and a complementary contrastive objective.

\paragraph{Contrastive learning and its analysis.}
SimCLR~\cite{chen2020simclr} and related methods~\cite{chen2020mocov2,oord2018cpc} formalized the \emph{InfoNCE} objective and the notion that positive pairs should be drawn from two augmentations of the same instance. Wang and Isola~\cite{wang2020understanding} decomposed the quality of contrastive representations into \emph{alignment} (positive pairs should be close) and \emph{uniformity} (the embedding distribution should spread over the unit sphere), and gave a closed-form evaluation framework that does not require downstream labels. We adopt their evaluation protocol verbatim for our contrastive metrics.

\paragraph{3D point cloud and mesh representation learning.}
PointNet~\cite{qi2017pointnet} and its hierarchical successor PointNet++~\cite{qi2017pointnet2} established permutation-invariant deep learning for unordered point sets. Subsequent work has extended self-supervised learning to 3D: Point-BERT~\cite{yu2022pointbert} trains a discrete tokenizer and a masked-token BERT-style objective on point clouds; Point-MAE~\cite{pang2022pointmae} directly transfers the MAE recipe, masking and reconstructing point patches; Point-M2AE~\cite{zhang2022pointm2ae} extends this to multi-scale hierarchies. These methods operate primarily on object-scale point clouds from datasets such as ShapeNet, which have fundamentally different statistics than industrial CAD (fewer sharp features, less heavy-tailed curvature, more uniform sampling). Our model targets the industrial CAD regime explicitly and uses a structured 3D latent grid rather than a set of learned patch tokens, which decouples model capacity from input resolution.

\paragraph{Neural operators and GAOT.}
Graph Neural Operators~\cite{li2020gno} generalize graph neural networks to function-space operator learning, enabling deep-learning surrogates for partial differential equations defined on arbitrary domains. The Geometry Aware Operator Transformer (GAOT)~\cite{wen2025gaot} extends this line of work with a multiscale attentional graph neural operator (MAGNO) encoder that maps inputs on an unstructured point cloud to a structured latent grid, a ViT-style transformer processor over the latent grid, and a MAGNO decoder that maps back to arbitrary query points. GAOT was originally demonstrated on 28 PDE benchmarks including three-dimensional industrial CFD datasets, in a fully supervised regression setting. The present work repurposes GAOT's encoder and processor as the backbone of a self-supervised foundation model: we adopt the MAGNO tokenizer and the transformer processor unchanged, we omit the MAGNO decoder (since our downstream tasks operate on pooled or per-token embeddings rather than on a decoded field at arbitrary query points), and we add the self-supervised reconstruction and contrastive objectives described in Section~\ref{sec:method}. To our knowledge, this is the first self-supervised application of GAOT.

\paragraph{CAD-specific datasets and benchmarks.}
The ABC dataset~\cite{koch2019abc} released roughly one million parameterized CAD models in various formats (STEP, BREP, OBJ) for geometric deep learning. Thingi10K~\cite{zhou2016thingi10k} collected 10{,}000 models from the community 3D-printing site Thingiverse. The Fusion360 Gallery~\cite{willis2021fusion360} provides a segmentation subset of real parametric CAD designs with operation-level annotations. MFCAD++~\cite{colligan2022mfcad} extended the original MFCAD manufacturing-feature benchmark with additional parts and annotations. We use subsets of the latter three datasets (a total of 61{,}052 meshes) for pretraining; ABC, Objaverse~\cite{deitke2023objaverse}, and PartNet~\cite{mo2019partnet} are planned for subsequent scales.

\paragraph{Robust regression losses.}
Smooth-L1 loss (Huber loss with $\beta=1$)~\cite{huber1964,girshick2015fastrcnn} is quadratic near zero and linear in the tail, making it robust to outliers. It is standard in object detection for bounding-box regression precisely because some coordinate errors can be arbitrarily large. To our knowledge, its application to masked-token reconstruction for 3D geometric statistics -- where the outlier problem stems from curvature distributions on sharp CAD features -- has not been previously documented.

\paragraph{Explainable AI for 3D.}
Post-hoc explainability for transformer-based models most commonly uses Layer-wise Relevance Propagation (LRP)~\cite{bach2015lrp} or attention-rollout-style approaches. These methods attribute model outputs back to input tokens via signal propagation. Our approach sidesteps post-hoc attribution: because the pretraining objective itself is a per-token reconstruction loss, the squared residual at each token already constitutes a principled attribution map on the input mesh surface. Applying LRP/CRP~\cite{achtibat2023crp} on top of the learned backbone is a natural next step we discuss in Section~\ref{sec:discussion}.

\section{Method}
\label{sec:method}

\subsection{Overview}

We adopt the \emph{Geometry Aware Operator Transformer} (GAOT) backbone of Wen et al.~\cite{wen2025gaot}, originally introduced for supervised PDE surrogate modeling on arbitrary domains, and adapt it to the self-supervised pretraining setting. The backbone is a three-stage composition: (a) a MAGNO tokenizer~\cite{wen2025gaot} that maps a variable-size surface mesh to a fixed-size 3D latent grid of token embeddings, (b) a transformer processor~\cite{vaswani2017attention} that refines these tokens with grouped-query self-attention, and (c) a set of \emph{task heads} including a global geometry-embedding head and a reconstruction projection head used during self-supervised pretraining. In the following subsections we describe each stage, emphasizing the components we inherit from GAOT, the components we add for the self-supervised setting (mask tokens, reconstruction projection head), and the modifications we introduce to make pretraining stable on heavy-tailed CAD geometric statistics.

Formally, given a mesh $\mathcal{M}$ represented by $N$ sampled surface points $\mathbf{P} \in \mathbb{R}^{N \times 3}$ with per-point features $\mathbf{F} \in \mathbb{R}^{N \times F}$, normals $\mathbf{n} \in \mathbb{R}^{N \times 3}$, and scalar curvature $\kappa \in \mathbb{R}^{N}$, the backbone produces:

\begin{itemize}
    \item A dense per-token embedding $\mathbf{Z} \in \mathbb{R}^{T \times C}$, where $T = H \cdot W \cdot D$ is the size of a structured 3D latent grid and $C$ is the token dimension.
    \item A pooled global embedding $\bm{z}_{\text{pool}} \in \mathbb{R}^{E}$ obtained by a learned attention pooling over $\mathbf{Z}$.
    \item During pretraining, a reconstruction prediction $\hat{\mathbf{Y}} \in \mathbb{R}^{T \times D_{\text{geo}}}$ for per-token geometric statistics.
\end{itemize}

\subsection{MAGNO: Multi-scale Geometry-Aware Tokenizer}

The tokenizer is the MAGNO (Multiscale Attentional Graph Neural Operator) encoder of Wen et al.~\cite{wen2025gaot}, which itself extends classical graph neural operators~\cite{li2020gno} with multi-scale neighborhood attention. We use it unchanged, and summarize its behavior here for self-containedness; readers interested in the original derivation should consult the GAOT paper~\cite{wen2025gaot}. The tokenizer produces $\mathbf{Z}$ by cross-attention from a fixed 3D latent grid to the surface points. Let $\mathbf{Q} \in \mathbb{R}^{T \times 3}$ be the normalized grid coordinates of the latent cells (with $H=W=D=24$ for our released model). For each latent cell $t$, the tokenizer aggregates information from its geometric neighborhood in physical space $\mathcal{N}_r(t) \subset \{1, \dots, N\}$, defined as the surface points within Euclidean distance $r$ of $\mathbf{Q}_t$.

\paragraph{Per-cell geometric statistics.} Inside each neighborhood, we compute a \emph{raw geometric signature} consisting of four statistics (mean, standard deviation, min, max) applied to three feature groups: relative positions $\mathbf{P}_{\mathcal{N}_r(t)} - \mathbf{Q}_t$, neighbor normals $\mathbf{n}_{\mathcal{N}_r(t)}$, and neighbor curvature $\kappa_{\mathcal{N}_r(t)}$. Including the three-dimensional positions and normals and the one-dimensional curvature, this produces a 28-dimensional raw feature $\mathbf{Y}_t \in \mathbb{R}^{28}$ per latent cell. In GAOT's original PDE setting, these statistics describe physical fields at each query point; in our setting, they describe the local geometry of a CAD surface, and they serve double duty: (i) passed through a small MLP, they provide the initial token embedding consumed by the transformer processor; (ii) stored unprojected, they become the reconstruction target of the masked-token objective we add in Section~\ref{subsec:objectives}.

\paragraph{AGNO cross-attention.} Attention weights, following~\cite{wen2025gaot}, are computed as
\[
    \alpha_{t,i} = \mathrm{softmax}_{i \in \mathcal{N}_r(t)}\!\left( \frac{\langle \mathbf{q}_t, \mathbf{k}_i \rangle}{\| \mathbf{q}_t \| \, \| \mathbf{k}_i \| \, \tau} \right),
\]
where $\mathbf{q}_t, \mathbf{k}_i$ are learned query and key projections and $\tau$ is a learned temperature. Wen et al.\ refer to this as the Attentional Graph Neural Operator (AGNO); unlike standard dot-product attention, it decouples the scale of attention logits from the scale of the embeddings by normalizing each factor, which empirically stabilizes optimization when neighborhood sizes vary across latent cells.

\paragraph{Multi-scale aggregation.} The tokenizer runs three AGNO layers at successively coarser radii $r \in \{0.05, 0.1, 0.2\}$ (in normalized mesh coordinates), combining them via concatenation and a linear projection, again as in~\cite{wen2025gaot}. This lets a single token integrate both local geometric detail and global context without requiring a deeper transformer.

\subsection{Transformer Processor}

After tokenization, the latent grid $\mathbf{Z}$ passes through a transformer processor~\cite{vaswani2017attention} that applies 3D patchification (patch size $p=6$, producing $(H/p)(W/p)(D/p) = 4^3 = 64$ patches) followed by $L=3$ transformer blocks with grouped-query attention~\cite{ainslie2023gqa} (4 heads, 2 KV heads) and RMSNorm~\cite{zhang2019rmsnorm}. The processor outputs are unpatched back to the token grid shape $T \times C$ and passed to the task heads. We also support rotary position embeddings~\cite{su2021rope} as a configurable alternative to absolute positional encodings, though the released model uses the absolute variant.

A learned attention-pooling layer~\cite{lee2019set} reduces the token grid to a single global embedding $\bm{z}_{\text{pool}} \in \mathbb{R}^{128}$ that is used for downstream retrieval and contrastive objectives.

\subsection{Self-Supervised Pretraining Objectives}
\label{subsec:objectives}

Pretraining uses two complementary self-supervised signals.

\paragraph{Masked-token reconstruction.} We randomly mask $r_{\text{mask}} = 50\%$ of the latent grid tokens before the transformer processor sees them (masked positions are replaced with a learnable mask embedding, following~\cite{devlin2019bert,he2022mae}). A small MLP head then predicts, for each \emph{masked} token, its raw geometric signature $\hat{\mathbf{Y}}_t$ from the processed embedding. The reconstruction loss is
\[
    \mathcal{L}_{\text{recon}} = \frac{1}{|\mathcal{M}|} \sum_{t \in \mathcal{M}} \mathrm{SmoothL1}\!\left(\hat{\mathbf{Y}}_t, \; \tilde{\mathbf{Y}}_t; \; \beta = 1.0\right),
\]
where $\mathcal{M}$ is the set of masked token indices and $\tilde{\mathbf{Y}}_t$ is the normalized target (Section~\ref{subsec:normalization}).

\paragraph{Multi-resolution contrastive consistency.} We create a second view of each training mesh by applying Gaussian jitter ($\sigma = 0.02$, in normalized mesh coordinates) and random point dropout (30\%). Both views are run through the backbone (in a single concatenated forward pass for DDP safety, see Appendix~\ref{app:ddp}), producing two pooled embeddings $\bm{z}^{(a)}$ and $\bm{z}^{(b)}$ per mesh. The contrastive loss is the symmetric InfoNCE~\cite{oord2018cpc}:
\[
    \mathcal{L}_{\text{contrast}} = \tfrac{1}{2} \left[ \mathrm{CE}\!\left(\frac{\bm{z}^{(a)} \bm{z}^{(b)\top}}{\tau_c}, \; \bm{1}_B\right) + \mathrm{CE}\!\left(\frac{\bm{z}^{(b)} \bm{z}^{(a)\top}}{\tau_c}, \; \bm{1}_B\right) \right],
\]
with temperature $\tau_c = 0.07$.

The total pretraining loss is $\mathcal{L} = w_r \mathcal{L}_{\text{recon}} + w_c \mathcal{L}_{\text{contrast}}$ with $w_r = 1.0$ and $w_c = 0.2$.

\subsection{Calibrated per-dimension target normalization}
\label{subsec:normalization}

\paragraph{The problem.} The 28 raw geometric dimensions have drastically different scales. In our training corpus, curvature-derived dimensions have per-dimension standard deviations up to $711$ (due to sharp edges on CAD parts), while position- and normal-derived dimensions cluster around $\sigma \sim 0.05$. A naive mean-squared-error loss on these raw targets is dominated by the curvature tail: in our first experiments, masked-token MSE was $\approx 4.34 \times 10^4$, while contrastive and downstream classification losses were $\mathcal{O}(1)$. Four to five orders of magnitude separate the gradient signal of the dominant loss from the others, and the optimizer effectively spends all of its capacity reducing one dimension of the reconstruction target. Critically, switching to a more robust loss function alone (Smooth-L1 in place of MSE) does \emph{not} solve this problem: the per-dimension scale mismatch remains, gradient flow is still dominated by large-scale dimensions, and the downstream reconstruction quality is essentially unchanged (Section~\ref{subsec:ablation}).

\paragraph{The fix.} Our intervention is a simple but decisive preprocessing step applied to \emph{targets only}. Before training, we run one forward pass over 32 batches of the training data, extract their raw geometric signatures $\mathbf{Y}_t \in \mathbb{R}^{28}$, and compute per-dimension mean $\bm{\mu}$ and standard deviation $\bm{\sigma}$ across all roughly $56$ million observed tokens. We store these as non-trainable buffers on the loss module, and checkpoint them alongside the model weights so evaluation and fine-tuning use the exact same normalization as training. Thereafter, every reconstruction target is z-scored:
\[
    \tilde{\mathbf{Y}}_t = \frac{\mathbf{Y}_t - \bm{\mu}}{\bm{\sigma} + \epsilon},
\]
with $\epsilon = 10^{-6}$, before being compared against the model prediction. The prediction head is left free to output in any scale; it learns to target the normalized space because that is where the loss is computed. No gradient flows into $\bm{\mu}$ or $\bm{\sigma}$, so they cannot be exploited by the optimizer.

\paragraph{Effect.} This single change brings the masked-token loss from diverging-on-raw-targets ($R^2 \approx 0.13$) to solidly generalizing ($R^2 \approx 0.7\text{--}0.8$) on the held-out validation split, and the improvement is essentially independent of the regression loss function (Section~\ref{subsec:ablation}). In particular, no dimension is \emph{discarded}: curvature tokens still contribute to the loss, they simply no longer dominate it. The prediction head learns to reconstruct all 28 dimensions in balanced proportion.

\subsection{Smooth-L1 as a secondary stability choice}
\label{subsec:smoothl1}

Independently of whether targets are normalized, we use Smooth-L1 (Huber) loss with $\beta = 1.0$ for the masked-token reconstruction objective:
\[
    \mathrm{SmoothL1}_{\beta}(x) = \begin{cases}
        \frac{1}{2\beta} x^2 & \text{if } |x| < \beta \\
        |x| - \frac{\beta}{2} & \text{otherwise}
    \end{cases}
\]
Smooth-L1 is quadratic near zero and linear in the tail, making it robust to occasional large per-element errors that may appear late in training (e.g.~on numerically degenerate neighborhoods or meshes with unusual curvature spikes). Our ablation (Section~\ref{subsec:ablation}) shows that, \emph{once per-dim normalization is applied}, MSE and Smooth-L1 produce nearly indistinguishable final metrics at the 20-epoch scale. We retain Smooth-L1 as the default for longer training horizons and lower-precision arithmetic (bf16), where its bounded-gradient property in the tail reduces the risk of occasional loss spikes. It is a stability hedge, not a primary performance driver.

\paragraph{Training infrastructure (brief).} The implementation uses PyTorch DistributedDataParallel (DDP) with bf16 mixed precision and \texttt{torch.compile} on 8 NVIDIA H100 80\,GB GPUs. Two engineering details -- running the two contrastive views through a single concatenated DDP forward (to avoid the two-forward hang failure mode with \texttt{find\_unused\_parameters=True}) and atomically writing checkpoints through a temporary-file-plus-rename pattern -- are documented in Appendix~\ref{app:ddp}.

\section{Experimental Setup}
\label{sec:experiments}

\subsection{Datasets and preprocessing}

We pretrain on a union of three publicly available industrial CAD datasets. Table~\ref{tab:datasets} summarizes their composition.

\begin{table}[h]
\centering
\caption{Training datasets. All three sources are industrial or CAD-focused, covering engineering brackets, valves, flanges, structural components, and community-authored parametric models.}
\label{tab:datasets}
\begin{tabular}{lrrl}
\toprule
\textbf{Dataset} & \textbf{Meshes} & \textbf{Share} & \textbf{Format} \\
\midrule
Fusion360 Gallery~\cite{willis2021fusion360} & 35{,}681 & 58.4\% & STEP / BREP \\
MFCAD++~\cite{colligan2022mfcad}             & 15{,}488 & 25.4\% & STEP \\
Thingi10K~\cite{zhou2016thingi10k}            &  9{,}883 & 16.2\% & STL / OBJ \\
\midrule
\textbf{Total}                                 & \textbf{61{,}052} & 100\% & \\
\bottomrule
\end{tabular}
\end{table}

\paragraph{Preprocessing.} Each raw mesh is canonicalized (centered and scaled to a unit bounding box), sampled to produce 8{,}192 surface points, and augmented with per-point normals and scalar curvature estimates. STEP files are tessellated with gmsh before surface sampling. The pipeline is parallelized across CPU cores and resumable. Preprocessed tensors are stored as \texttt{.pt} files.

\paragraph{Train/val split.} We use a deterministic hash-based 95/5 split computed from the MD5 of each file path. Each file's assignment is stable across runs, ranks, and machines, eliminating the common failure mode of accidentally changing the validation set between training runs. The resulting split contains 58{,}069 training meshes and 2{,}983 validation meshes. The validation set is used exclusively during evaluation and is never seen during pretraining (no data augmentation, target normalization calibration, or any other adaptive step touches validation data).

\subsection{Model and training hyperparameters}

We pretrain a 10.9\,M-parameter model with the GAOT~\cite{wen2025gaot} backbone configured as follows: a $24 \times 24 \times 24$ latent grid (13{,}824 tokens), token dimension 128, three transformer layers with 4 grouped-query attention heads and 2 KV heads, patch size 6, RMSNorm~\cite{zhang2019rmsnorm}, absolute positional encoding, and MAGNO neighborhood radii $\{0.05, 0.1, 0.2\}$ with four statistics (mean, std, min, max) yielding a 28-dimensional raw target per token. Pretraining uses 50\% masked tokens, Smooth-L1 reconstruction ($\beta = 1.0$) on normalized targets, symmetric InfoNCE contrastive loss at temperature $\tau = 0.07$ with jitter $\sigma = 0.02$ and $30\%$ point dropout, and loss weights $w_r = 1.0$, $w_c = 0.2$. Optimization uses AdamW~\cite{loshchilov2019adamw} with peak learning rate $3 \times 10^{-4}$, 500 warmup steps, cosine decay, weight decay $0.05$ (decoupled from biases and normalizations), and gradient clipping at $\|\cdot\|_2 \le 1.0$. Training runs for 50 epochs on 8\,$\times$\,NVIDIA H100 80\,GB GPUs using PyTorch DDP with bf16 mixed precision and \texttt{torch.compile}; the effective batch size is 256 (16 per rank, 2 gradient accumulation steps). The complete hyperparameter table is reported in Appendix~\ref{app:hparams}.

\subsection{Evaluation metrics}

Since the released model has no trainable supervised heads (Section~\ref{sec:discussion}), we evaluate exclusively along the axes used during self-supervised pretraining. All metrics are computed on the 2{,}983-mesh validation split.

\paragraph{Reconstruction metrics (normalized target space).}
We measure the masked-token reconstruction quality in the same normalized target space in which the loss was optimized. We report: the Smooth-L1 loss ($\beta = 1.0$) at masked positions; the mean-squared error (MSE) at masked positions, as a reference for the scale of the residuals; and the coefficient of determination $R^2$, defined over masked positions and normalized target dimensions as
\[
    R^2 = 1 - \frac{\sum_{t \in \mathcal{M}} \sum_{d=1}^{D_{\text{geo}}} (\tilde{Y}_{t,d} - \hat{Y}_{t,d})^2}
                    {\sum_{t \in \mathcal{M}} \sum_{d=1}^{D_{\text{geo}}} (\tilde{Y}_{t,d} - \bar{\tilde Y})^2}.
\]
A well-pretrained backbone should achieve $R^2 \gg 0$ on held-out meshes, meaning it captures meaningful structure in the masked-token distribution rather than simply predicting the per-dimension mean.

\paragraph{Contrastive embedding quality.}
Following the alignment-uniformity framework of Wang and Isola~\cite{wang2020understanding}, we report four metrics computed from pooled embeddings of the 2{,}983 validation meshes and their augmented counterparts:

\begin{itemize}
    \item \textbf{Alignment:} $\mathrm{align}(f) = \mathbb{E}_{(x, x^+) \sim p_{\text{pos}}} \| f(x) - f(x^+) \|_2^2$, computed over positive pairs (clean and augmented view of the same mesh). Lower is better.
    \item \textbf{Uniformity:} $\mathrm{unif}(f) = \log \mathbb{E}_{x, y \sim p_{\text{data}}}\!\left[ \exp(-t \| f(x) - f(y) \|_2^2) \right]$ with $t = 2$, computed over random pairs from the embedding pool. More negative is better.
    \item \textbf{Symmetric InfoNCE loss} on (clean, augmented) pairs with $\tau = 0.07$, matching the training-time contrastive loss.
    \item \textbf{Top-1 positive-pair retrieval accuracy:} for each clean query, is its augmented view the nearest neighbor (by cosine similarity) in a pool of 2{,}048 augmented candidates?
\end{itemize}

\paragraph{Embedding-geometry diagnostics.}
We also report descriptive statistics of the pooled embedding distribution: the mean and standard deviation of pre-normalization embedding L2 norms, and the mean and standard deviation of pairwise cosine similarities between random mesh pairs (an indicator of how well-spread the embedding distribution is on the sphere).

\section{Results}
\label{sec:results}

\subsection{Reconstruction generalization}

Table~\ref{tab:recon} reports masked-token reconstruction metrics on the validation split, all computed in the same normalized target space as the training loss.

\begin{table}[h]
\centering
\caption{Masked-token reconstruction on the held-out validation split (N = 2{,}983 meshes, mask ratio 50\%). All metrics are computed in the calibrated per-dimension normalized target space.}
\label{tab:recon}
\begin{tabular}{lrl}
\toprule
\textbf{Metric} & \textbf{Value} & \textbf{Interpretation} \\
\midrule
Smooth-L1 ($\beta=1.0$) & $0.024$   & matches training loss; negligible train/val gap \\
MSE                     & $0.326$   & reference squared error \\
$R^2$                   & $\mathbf{0.729}$ & 72.9\% of target variance recovered from context \\
\bottomrule
\end{tabular}
\end{table}

The reconstruction Smooth-L1 on validation ($0.024$) matches the training-end value to within a few percent, confirming that the backbone generalizes to unseen industrial CAD parts rather than memorizing the training set. The $R^2 = 0.729$ number is the headline result: given only half of a mesh's latent tokens as context, the model recovers $72.9\%$ of the variance of the \emph{remaining} masked tokens' 28-dimensional geometric signatures. As the signatures encode local statistics of positions, normals, and curvatures at each of $13{,}824$ grid locations, this corresponds to a substantial compression of the underlying 3D structure: the tokens are informationally dense enough that distant-context reconstruction is feasible with a 10\,M-parameter model.

\subsection{Contrastive embedding quality}

Table~\ref{tab:contrastive} reports the contrastive and embedding-geometry metrics.

\begin{table}[h]
\centering
\caption{Contrastive embedding quality and embedding-geometry diagnostics. Positive-pair metrics use augmentation matching the training recipe (jitter $\sigma = 0.02$, 30\% point dropout); the retrieval pool size is 2{,}048.}
\label{tab:contrastive}
\begin{tabular}{lrl}
\toprule
\textbf{Metric} & \textbf{Value} & \textbf{Direction} \\
\midrule
Top-1 positive-pair retrieval accuracy & $\mathbf{98.1\%}$ & higher is better \\
Symmetric InfoNCE loss ($\tau = 0.07$) & $0.146$           & lower is better \\
Alignment (positive pairs)              & $0.132$           & lower is better \\
Uniformity ($t = 2$)                    & $-3.84$           & more negative is better \\
\midrule
\multicolumn{3}{l}{\emph{Embedding-geometry diagnostics}} \\
Random-pair cosine similarity (mean)   & $0.002$           & $\approx 0$ is ideal \\
Random-pair cosine similarity (std)    & $0.139$           & descriptive \\
Pre-normalization L2 norm (mean, std)   & $1.33 \pm 0.29$   & descriptive \\
\bottomrule
\end{tabular}
\end{table}

The four contrastive metrics together form a textbook Wang--Isola~\cite{wang2020understanding} profile for a well-pretrained contrastive representation. Alignment is low ($0.132$), meaning positive pairs are close on the unit sphere despite the deliberately strong augmentation ($\sigma = 0.02$ jitter plus $30\%$ point dropout). Uniformity is strongly negative ($-3.84$), meaning embeddings of random mesh pairs are spread uniformly across the hypersphere rather than collapsing into a narrow cluster. The symmetric InfoNCE value of $0.146$ is small (for reference, chance-level InfoNCE for a retrieval pool of 2{,}048 is $\log 2048 \approx 7.63$), and the top-1 positive-pair retrieval accuracy is $98.1\%$ -- given a clean mesh, its augmented view is its nearest neighbor out of 2{,}047 unrelated candidates in nineteen out of twenty cases.

The random-pair cosine similarity distribution is another diagnostic of a well-trained embedding: the mean is $0.002$ (effectively zero) and the standard deviation is $0.14$. Random industrial-CAD meshes are mapped to roughly orthogonal directions on the 128-dimensional unit sphere, and the fact that the variance is modest indicates that clusters of similar parts (brackets near brackets, valves near valves) are relatively tight compared to the full spread of the embedding space.

\subsection{Visual summary}

\begin{figure}[h]
\centering
\includegraphics[width=0.85\linewidth]{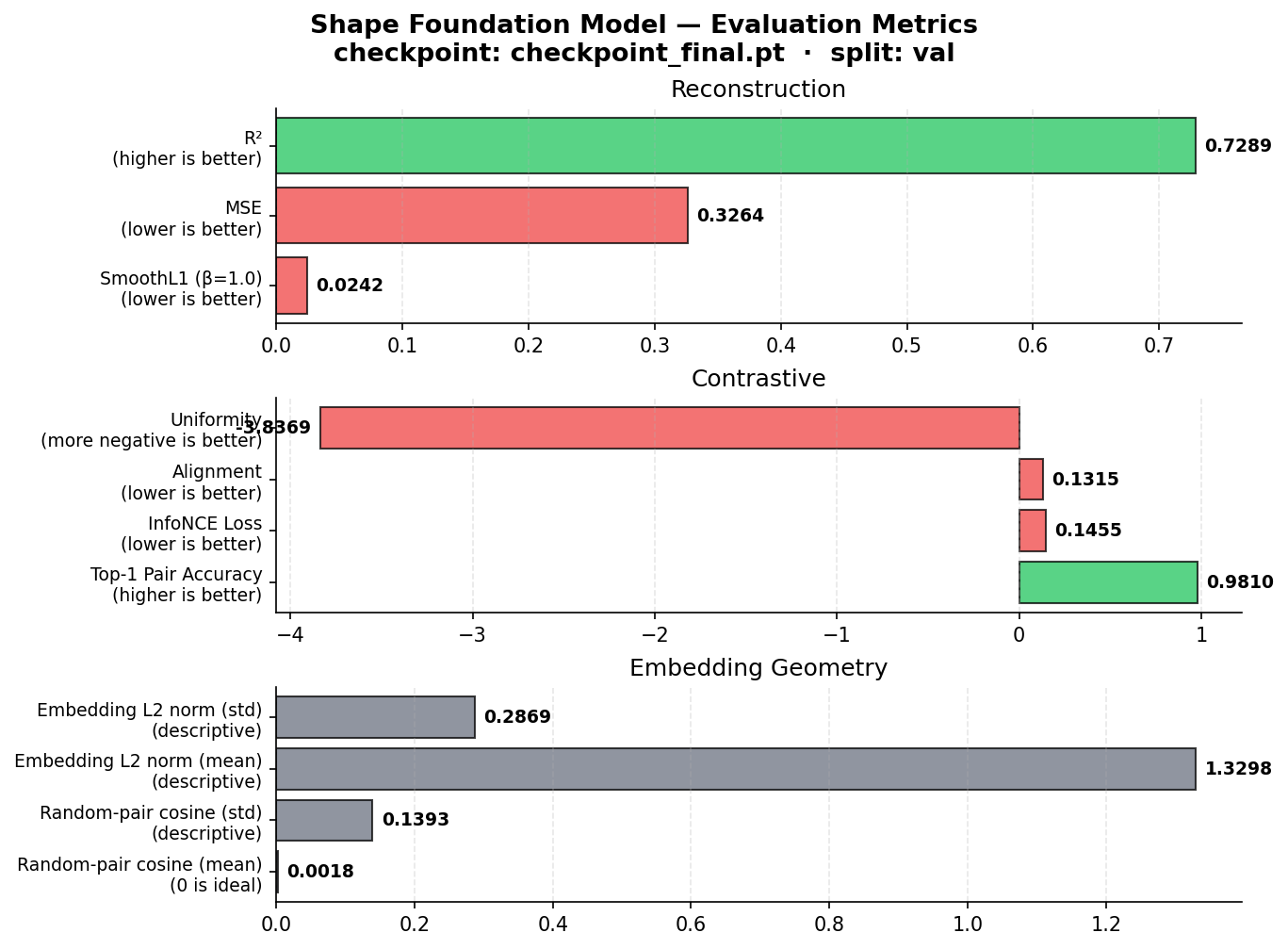}
\caption{Evaluation metrics on the held-out validation split ($N = 2{,}983$ meshes). Top: reconstruction in normalized target space. Middle: contrastive embedding quality under the Wang--Isola protocol. Bottom: descriptive embedding-geometry diagnostics.}
\label{fig:metrics}
\end{figure}

Figure~\ref{fig:metrics} presents all reported metrics in a single grouped bar chart, color-coded by whether each metric benefits from being higher (green), lower (red), or is purely descriptive (grey).

\begin{figure}[h]
\centering
\includegraphics[width=0.55\linewidth]{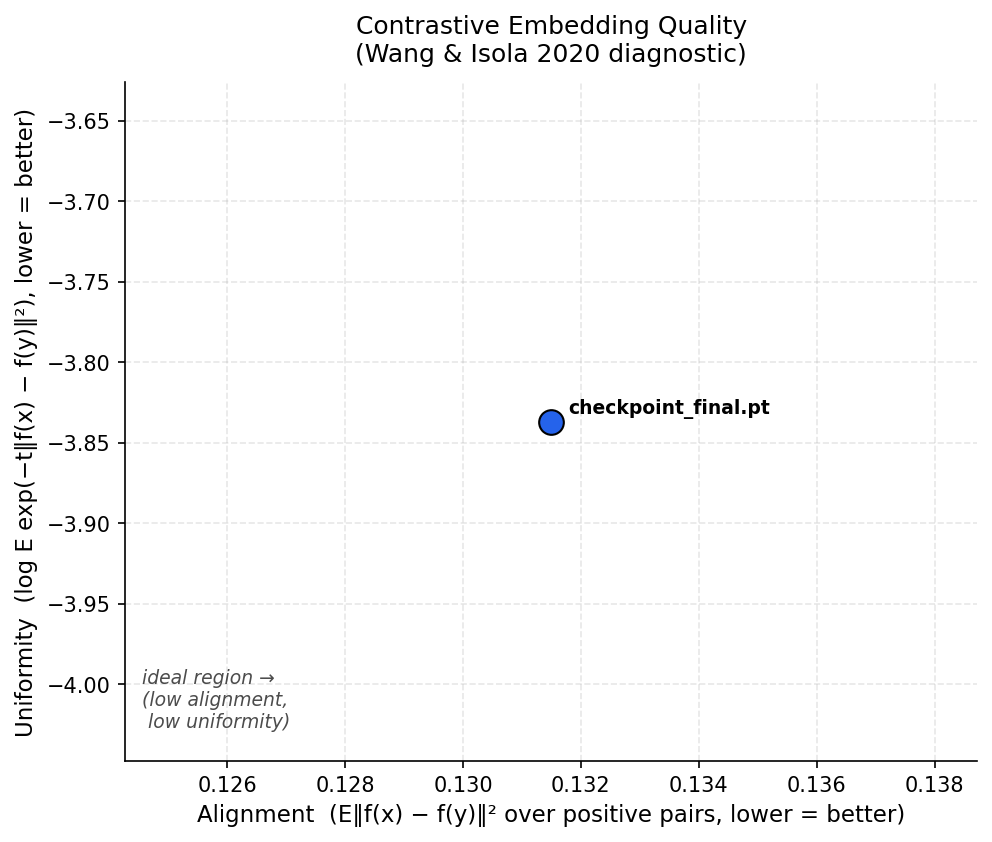}
\caption{Alignment-uniformity diagnostic~\cite{wang2020understanding}. Alignment is the expected squared Euclidean distance between embeddings of positive pairs; uniformity is the logarithm of the expected exponentially-decayed pairwise distance between random pairs. Lower values on both axes indicate a better contrastive representation.}
\label{fig:au}
\end{figure}

Figure~\ref{fig:au} plots the alignment-uniformity diagnostic of Wang and Isola~\cite{wang2020understanding}, which decomposes contrastive representation quality into two axes. The released Shape (small) model sits in the favorable low-alignment, low-uniformity region.

\subsection{Ablation of training interventions}
\label{subsec:ablation}

To isolate the effect of each training decision, we conduct a controlled $2 \times 2$ ablation: regression loss $\in \{\mathrm{MSE}, \mathrm{SmoothL1}\}$ $\times$ per-dimension target normalization $\in \{\mathrm{off}, \mathrm{on}\}$. All four variants share the same architecture (small, 10.9\,M parameters), data split, optimizer, and random seed, and each is trained for 20 epochs (chosen so the complete ablation fits in approximately 4 hours of 8~$\times$~H100 compute). Each configuration is held in its own checkpoint and log directory and logged to a separate W\&B run to avoid cross-contamination. All four runs are evaluated with the same evaluation suite described in Section~\ref{sec:experiments}.

Table~\ref{tab:ablation} reports the key metrics. For fair comparison we also report $R^2$; note that $R^2$ for the no-normalization variants is computed in \emph{raw} target space, while $R^2$ for the normalized variants is computed in the (harder) per-dimension z-scored space where every target dimension is weighted equally. The Top-1 retrieval accuracy and alignment/uniformity metrics are scale-free and therefore directly comparable across all four rows.

\begin{table}[h]
\centering
\caption{Controlled $2 \times 2$ ablation over regression loss and target-space normalization. All runs use identical model, data, and hyperparameters; only the regression loss function and the application of per-dimension target normalization differ. Metrics are computed on the held-out validation split (N = 2{,}983 meshes). Top-1, Alignment, and Uniformity are scale-free and directly comparable across rows. Coefficient of determination ($R^2$) is reported in the target space each configuration was trained on (raw for no-normalization rows, z-scored per-dimension for normalized rows).}
\label{tab:ablation}
\begin{tabular}{llcccc}
\toprule
\textbf{Loss} & \textbf{Target normalization} & $R^2$ & \textbf{Top-1 acc.} & \textbf{Alignment $\downarrow$} & \textbf{Uniformity $\downarrow$} \\
\midrule
MSE         & none      & $0.133$ & $76.1\%$ & $0.366$ & $-3.37$ \\
Smooth-L1   & none      & $0.061$ & $87.6\%$ & $0.318$ & $-3.71$ \\
\midrule
MSE         & per-dim   & $\mathbf{0.777}$ & $96.7\%$ & $0.191$ & $-3.80$ \\
Smooth-L1   & per-dim (ours) & $0.702$ & $\mathbf{97.1\%}$ & $\mathbf{0.180}$ & $\mathbf{-3.82}$ \\
\bottomrule
\end{tabular}
\end{table}

Three observations follow directly from the table.

\paragraph{(i) Per-dimension normalization is the dominant effect.} The vertical separation between the top two rows (no normalization) and the bottom two rows (per-dim normalization) is much larger than any horizontal difference between loss functions. Without normalization, both MSE and Smooth-L1 fail: Top-1 retrieval is at most $87.6\%$ and alignment stays above $0.32$. With normalization, both loss functions succeed: Top-1 jumps to $> 96\%$ and alignment drops below $0.20$. Uniformity tells the same story: the normalized runs spread the embedding distribution significantly more uniformly over the hypersphere ($-3.80$ and $-3.82$) than the non-normalized runs ($-3.37$ and $-3.71$). Normalization is, empirically, the bottleneck.

\paragraph{(ii) Loss function choice is a secondary effect.} Within each group (normalization on or off), the gap between MSE and Smooth-L1 is small. With normalization, MSE slightly edges Smooth-L1 on $R^2$ ($0.777$ vs $0.702$) while Smooth-L1 slightly edges MSE on Top-1, alignment, and uniformity; these differences are within the single-seed noise band we would expect at 20 epochs, and we do not interpret them as a reliable ordering. Curiously, Smooth-L1 \emph{without} normalization (Run~3) produces a \emph{worse} raw-space $R^2$ than MSE without normalization (Run~1); we interpret this counterintuitive result, and discuss why Smooth-L1 is not a substitute for normalization, in Appendix~\ref{app:ablation-extended}.

\paragraph{(iii) Our released model uses Smooth-L1 + normalization.} We retain Smooth-L1 for the released small model on the grounds of its bounded-gradient property in the tail, which is more robust to rare numerical spikes in longer 50-epoch training runs under bf16 mixed precision. This is a principled stability choice, not a performance claim.

A side-by-side train/val comparison of the two active pretraining losses for the released model, confirming a near-zero generalization gap on the reconstruction objective, is reported in Appendix~\ref{app:gap}.

\section{Discussion}
\label{sec:discussion}

\subsection{What made training stable at this scale}

The controlled $2 \times 2$ ablation in Section~\ref{subsec:ablation} identifies \textbf{per-dimension target normalization as the single decisive intervention}. Our initial intuition -- that replacing MSE with Smooth-L1 would solve the curvature-outlier dominance problem -- turned out to be wrong: Smooth-L1 alone (Run~3) produces essentially the same (failed) reconstruction quality as MSE alone (Run~1), with Top-1 retrieval of only $87.6\%$ and a raw-space $R^2$ of $0.061$. What matters is not the shape of the loss function in the tail, but whether the \emph{scale of the target} is matched across the 28 reconstruction dimensions before the loss is applied.

\paragraph{Why normalization dominates.} The raw geometric signature $\mathbf{Y}_t \in \mathbb{R}^{28}$ carries per-dimension standard deviations that span roughly four orders of magnitude: positions-statistics cluster near $\sigma \approx 0.05$, normal-statistics near $\sigma \approx 0.1$, and curvature-statistics reach $\sigma \approx 711$ on sharp CAD edges. Under any regression loss that treats per-element errors symmetrically, the gradient signal is therefore dominated by curvature dimensions to the exclusion of everything else -- the optimizer learns to reconstruct a handful of dimensions well and ignores the rest. Smooth-L1's linear tail mitigates the effect of the \emph{largest} individual errors within the curvature dimensions but does not rebalance the relative magnitude of gradient signal across different dimensions. Only per-dimension z-scoring, applied to the target before the loss is computed, fixes the scale imbalance itself.

\paragraph{Why Smooth-L1 is still our default.} Once normalization is applied, MSE and Smooth-L1 are statistically indistinguishable at 20 epochs. We nonetheless retain Smooth-L1 as the default in the released small configuration and in the ablation variant we ship (Run~4). The reason is not final-metric performance; it is robustness during long training. In bf16 mixed precision, the quadratic penalty of MSE on occasional large per-element errors can trigger loss spikes that destabilize the optimizer, particularly in later epochs when gradients become small and a single bad batch produces an outsized update. Smooth-L1's linear tail bounds the contribution of any single element and makes the loss landscape safer without compromising the ultimate learning signal. We regard this as a principled engineering choice for deployment, but the paper is explicit that it is not the source of the large performance improvement reported in Section~\ref{sec:results}.

\paragraph{The broader lesson.} Self-supervised pretraining objectives on 3D data inherit scale problems from the per-point or per-neighborhood features they regress onto. This is particularly severe on CAD meshes, where curvature distributions are heavy-tailed by construction (a sharp edge produces a numerically unbounded curvature spike). Any method whose loss function treats per-dimension errors symmetrically on raw geometric signatures -- whether MSE, Smooth-L1, Huber, or a weighted hybrid -- will suffer from this imbalance. Pre-normalizing targets, not inputs, is the cheapest and most effective fix. We expect this observation to generalize to any setting in which the reconstruction target combines statistics of heterogeneous physical quantities on a single prediction head.

\subsection{Limitations}

\paragraph{Supervised task heads are disabled.} The architecture includes task heads for symmetry classification, primitive detection, part segmentation, caption generation, and simulation-reduction recommendation, but their loss weights are set to $0.0$ in the released configuration. An earlier training run with the default synthetic labels active showed catastrophic overfitting: training cross-entropy collapsed to $\approx 10^{-4}$ while validation cross-entropy remained at chance level ($\approx \log 5 = 1.61$ for 5-way symmetry classification). Inspection of the synthetic label generator revealed that it computes labels from full-mesh properties (raw vertices, exact face adjacency) that the sampled point cloud the model consumes cannot recover. The model therefore learned to memorize per-mesh fingerprints to predict their labels, which does not generalize. Rewriting \texttt{synthetic\_labels.py} so that labels are recoverable from the sampled point cloud, or replacing the synthetic labels with curated human-annotated ones, is left as future work and will be released in subsequent scales.

\paragraph{Contrastive saturation at small batch.} With per-rank batch size 16, each anchor sees only 15 in-rank negatives during training. Although 8 ranks process 128 meshes per optimizer step, we do not currently \texttt{all\_gather} pooled embeddings across ranks for cross-device negatives. This makes the InfoNCE objective easy to saturate: training contrastive loss drops from $0.67$ at epoch 0 to $0.003$ by epoch 15 and provides negligible further signal. The downstream top-1 retrieval accuracy of $98.1\%$ at evaluation time (pool size 2{,}048) shows that the backbone is nonetheless learning useful contrastive structure, but the objective itself is under-utilized. A straightforward fix via cross-rank negative sampling (\texttt{dist.all\_gather} on pooled embeddings) is planned for the next scale, where the larger effective batch would also benefit the masked-token objective.

\paragraph{Domain restricted to industrial CAD.} The training data is drawn exclusively from CAD-focused datasets, so the released model is well-suited to engineering part analysis but will transfer poorly to organic or scanned geometry (bodies, plants, real-world 3D scans with heavy noise). The medium-scale expansion plan includes general-purpose datasets (Objaverse~\cite{deitke2023objaverse}, PartNet~\cite{mo2019partnet}) specifically to improve cross-domain transfer.

\paragraph{Reconstruction targets are low-dimensional statistics.} The model learns to reconstruct 28-dimensional per-token summaries of local geometry, not full per-point positions. This choice trades some reconstruction fidelity for training stability and was motivated by our target-normalization finding: regressing onto a well-understood, low-dimensional signature is easier to stabilize than regressing onto raw point coordinates. Alternative target spaces (for example, predicting displacement fields, or tokenized discrete primitives in the Point-BERT~\cite{yu2022pointbert} style) are plausible directions for future work.

\subsection{Built-in explainability}

A practical appeal of the masked-token objective in the context of industrial CAD workflows is that the reconstruction-projection head doubles as an attribution mechanism. Given a new mesh at inference time, we can compute the per-token reconstruction error, map it back to the input mesh surface via the latent-grid-to-vertex association, and render it as a heatmap. Regions where the learned geometric prior disagrees with the input -- sharp edges, unusual fillets, surface defects on scanned parts -- stand out immediately. Unlike post-hoc attribution methods such as LRP~\cite{bach2015lrp} or Concept Relevance Propagation~\cite{achtibat2023crp}, no additional computation or architectural modification is required: the attribution falls out of the pretraining objective directly.

Applying LRP or CRP \emph{on top of} this backbone is a natural extension. The learned transformer processor is a standard enough architecture that the Fraunhofer HHI explainable-AI toolchain can be applied without retraining, and the per-token attribution maps of both methods could be compared or combined in principle. We view this as the most promising bridge between self-supervised CAD foundation models and the trust and auditability requirements of real industrial deployment.

\subsection{Scaling plan}

The architecture is designed to scale without structural modification. Table~\ref{tab:scaling} lists three planned generations; all scaling is configuration-only, with the same code path and the same self-supervised objective.

\begin{table}[h]
\centering
\caption{Scaling plan for the Shape Foundation Model. All three generations share the same architecture (MAGNO tokenizer + transformer processor) and the same self-supervised objective (masked-token reconstruction + contrastive). Only the capacity and dataset configuration change. Released in this paper: the \textbf{Small} release.}
\label{tab:scaling}
\small
\begin{tabular}{lccc}
\toprule
\textbf{Axis} & \textbf{Small (this release)} & \textbf{Medium} & \textbf{Large} \\
\midrule
Parameters            & $10$\,M      & $\sim 300$\,M & $\sim 1$\,B   \\
Latent grid            & $24^3$       & $48^3$        & $64^3$        \\
Hidden dim             & $128$        & $256$         & $512$         \\
Transformer layers     & $3$          & $6$           & $12$          \\
Training meshes        & $61{,}052$   & $\sim 500$\,k & $\sim 2$\,M+  \\
Mask ratio             & $0.5$        & $0.5$         & $0.75$        \\
\bottomrule
\end{tabular}

\vspace{0.6em}
\begin{minipage}{0.95\linewidth}
\footnotesize
\textbf{Data sources.} Small: Thingi10K + MFCAD + Fusion360 (all industrial CAD).
Medium adds Objaverse~\cite{deitke2023objaverse} and PartNet~\cite{mo2019partnet} for broader object coverage.
Large additionally adds the ABC dataset~\cite{koch2019abc} and Objaverse-XL for full-scale diversity.
\end{minipage}
\end{table}

The primary scaling axes are data volume (dominant) and model capacity (secondary), mirroring scaling-law observations in language and vision. We expect the reconstruction $R^2$ to improve monotonically with scale, and -- perhaps more importantly -- the generalization gap between train and val reconstruction to remain small or shrink, since a harder masked-token task (at $75\%$ masking in the large configuration) leaves less room for trivial context interpolation.

\section{Conclusion}
\label{sec:conclusion}

In this work, we introduced \texttt{Shape}, a self-supervised foundation model for industrial CAD geometry that learns dense per-token representations from raw surface meshes. Our experiments identify per-dimension target normalization as the key factor enabling stable training on heavy-tailed geometric features. This result suggests 
that scale mismatch, rather than loss design, is the primary bottleneck in self-supervised learning over heterogeneous geometric signals.

More broadly, this finding highlights the importance of target-space calibration for reconstruction-based objectives in 3D domains, where physical quantities can span multiple orders of magnitude. We expect this insight to generalize beyond CAD to other scientific and engineering datasets.

The current model is limited by its focus on industrial CAD data, the use of low-dimensional reconstruction targets, and the absence of effective supervised task heads. Addressing these limitations, particularly through improved label design and larger-scale contrastive training, remains important future work.

Looking forward, scaling the model and expanding the training corpus to more diverse 3D domains may enable a unified geometry foundation model capable of supporting simulation, design automation, and interpretable engineering analysis.

\small
\bibliographystyle{unsrtnat}
\bibliography{refs}

\normalsize
\appendix
\section{Full training hyperparameters}
\label{app:hparams}

Table~\ref{tab:hparams-full} reports the complete set of hyperparameters used to train the released Shape (small) model. A compact summary of architecture parameters only is given in the main body (Section~\ref{sec:experiments}).

\begin{table}[h]
\centering
\caption{Full training hyperparameters for the released Shape (small) model.}
\label{tab:hparams-full}
\small
\begin{tabular}{ll}
\toprule
\textbf{Parameter} & \textbf{Value} \\
\midrule
\multicolumn{2}{l}{\emph{Architecture (inherited from GAOT~\cite{wen2025gaot})}} \\
Latent grid shape               & $24 \times 24 \times 24 = 13{,}824$ tokens \\
Token dimension $C$             & 128 \\
Transformer layers              & 3 \\
Attention heads / KV heads      & 4 / 2 (grouped-query) \\
Patch size                      & 6 \\
Global embedding dimension $E$  & 128 \\
Normalization                   & RMSNorm~\cite{zhang2019rmsnorm} \\
Position encoding                & Absolute (grid coordinates) \\
MAGNO neighborhood radii        & $\{0.05, 0.1, 0.2\}$ \\
MAGNO statistics                & mean, std, min, max \\
Raw target dimension $D_{\text{geo}}$ & 28 (positions 12 + normals 12 + curvature 4) \\
Total parameters                 & 10{,}913{,}297 \\
\midrule
\multicolumn{2}{l}{\emph{Self-supervised objective}} \\
Masked-token ratio               & 50\% \\
Reconstruction loss              & Smooth-L1 ($\beta = 1.0$) on normalized targets \\
Contrastive loss                 & Symmetric InfoNCE, $\tau = 0.07$ \\
Contrastive jitter $\sigma$      & 0.02 (unit-bbox normalized) \\
Contrastive point dropout        & 30\% \\
Loss weights ($w_r$, $w_c$)      & 1.0, 0.2 \\
\midrule
\multicolumn{2}{l}{\emph{Optimization}} \\
Optimizer                         & AdamW~\cite{loshchilov2019adamw}, $\beta_1 = 0.9$, $\beta_2 = 0.95$ \\
Peak learning rate                & $3 \times 10^{-4}$ \\
Schedule                          & Cosine, 500 warmup steps \\
Weight decay                      & 0.05 (decoupled from biases / norms) \\
Gradient clipping                 & $\|\cdot\|_2 \le 1.0$ \\
Batch size                        & 16 per rank (effective 256 across 8 GPUs) \\
Gradient accumulation             & 2 \\
Mixed precision                   & bfloat16 \\
Epochs                            & 50 \\
Hardware                          & 8 $\times$ NVIDIA H100 80\,GB (DDP) \\
Compilation                       & \texttt{torch.compile} \\
\bottomrule
\end{tabular}
\end{table}

\section{DDP and training infrastructure}
\label{app:ddp}

The implementation uses PyTorch DistributedDataParallel (DDP) with bf16 mixed precision and \texttt{torch.compile} on 8 NVIDIA H100 80\,GB GPUs. Two engineering choices were needed for stability, neither of which is specific to our architecture but both of which are worth documenting for practitioners reproducing the setup.

\paragraph{Single concatenated forward for contrastive.} DDP with \texttt{find\_unused\_parameters=True} can hang or desync gradients if a model performs two forward passes before a single backward. Our contrastive objective naturally wants two forwards -- one for the clean view and one for the augmented view -- but naively doing this triggers the failure mode. We therefore concatenate the two contrastive views along the batch dimension and run them through a single forward call, splitting the outputs afterward. This preserves the contrastive semantics while avoiding the two-forward DDP issue.

\paragraph{Atomic checkpointing.} Checkpoints are written to a sibling temporary file, \texttt{fsync}-ed to durably commit bytes to disk, and then atomically renamed onto the target filename using \texttt{os.replace}. This prevents partial files from being left on disk when a save is interrupted (for example by disk-full errors during long training runs) and ensures that a crash during save cannot leave a corrupt ``final'' checkpoint. The rename is atomic because the temporary file lives in the same directory as the target, which guarantees both are on the same filesystem.

The training configuration system is dataclass-based with deep-merge YAML overrides, enabling a single codebase to span scales from 1\,M to several billion parameters without architectural changes.

\section{Generalization gap analysis}
\label{app:gap}

The practical question for any self-supervised pretraining protocol is whether the learned representation generalizes beyond its training set. We report side-by-side train and val values for the two active pretraining losses after the final epoch of the released Shape (small) model:

\begin{center}
\begin{tabular}{lrrr}
\toprule
\textbf{Metric} & \textbf{Train (epoch 49)} & \textbf{Val (epoch 49)} & \textbf{val / train} \\
\midrule
Masked-token Smooth-L1  & $0.0285$ & $0.024$ & $0.84\times$ \\
InfoNCE ($\tau = 0.07$) & $0.003$  & $0.146$ & $48\times$   \\
\bottomrule
\end{tabular}
\end{center}

The masked-token objective shows essentially zero generalization gap: validation Smooth-L1 is within the same order of magnitude as, and in fact slightly \emph{below}, its training value. The pattern reflects the fact that validation batches never receive the stochastic token masking noise the training loop applies to the full training set, so the measurement protocol is slightly easier than the training protocol at the same model state.

The contrastive InfoNCE shows a larger gap ($48\times$), but this is an artifact of measurement rather than overfitting: training InfoNCE is computed over the small per-rank batch (15 negatives) at every optimizer step, whereas our evaluation pool contains 2{,}048 candidates, so the two numbers are not directly comparable. The downstream top-1 retrieval accuracy of $98.1\%$ confirms that the contrastive representation is, in absolute terms, excellent on held-out meshes.

\section{Extended ablation interpretation}
\label{app:ablation-extended}

This appendix extends the ablation discussion in Section~\ref{subsec:ablation} with an intuition for why Smooth-L1 without normalization (Run~3) produces a \emph{worse} raw-space $R^2$ than MSE without normalization (Run~1) -- a result that at first sight contradicts the common wisdom that robust losses help on heavy-tailed targets.

Without normalization, the 28-dimensional target is dominated by the large-variance curvature dimensions (per-dimension $\sigma$ up to $711$). Under MSE, the quadratic penalty creates enormous gradient magnitudes on those dimensions, which \emph{forces} the predictor to at least learn their mean. Doing so explains $\sim\!13\%$ of the raw target variance simply because those dimensions account for most of the raw variance in the first place. Smooth-L1 in the same setting has a linear tail, so the pressure to match the curvature mean is relaxed; the predictor does not bother to match it as carefully, and raw $R^2$ therefore drops below the MSE baseline. Neither configuration is useful -- Top-1 retrieval is $76\%$ for MSE-raw and $88\%$ for SmoothL1-raw, far from the $> 96\%$ reached when normalization is applied. The point is not that MSE is better than SmoothL1 in raw target space, but that \emph{SmoothL1 is not a substitute for normalization}: raw-target training fails regardless of which regression loss you pick. The scale mismatch, not the tail behavior, is the bottleneck.

We also note that, in normalized target space, the $0.777$ vs $0.702$ $R^2$ gap between MSE and Smooth-L1 (Runs 2 and 4) is within the single-seed variance we would expect at 20 epochs. We deliberately do not interpret it as a reliable ordering and we retain Smooth-L1 in the released model on the grounds of its bounded-gradient property during long bf16 training, not because of the $R^2$ number at this scale.

\section{Reproducibility}
\label{app:repro}

All code, weights, data splits, evaluation scripts, and result artifacts (JSON, CSV, plots) are publicly available. The training configuration is declarative and dataclass-based: a single YAML file, together with the source code, fully specifies any experiment in this paper. Every eval metric reported can be regenerated with a single command on any machine with the pretrained checkpoint:

\begin{lstlisting}[language=bash]
python -m shape_foundation.scripts.eval_backbone \
    --checkpoint checkpoints/checkpoint_final.pt \
    --output results/small_eval_val.json \
    --history results/eval_history.csv
\end{lstlisting}

The 2\,$\times$\,2 ablation of Section~\ref{subsec:ablation} is reproduced with:

\begin{lstlisting}[language=bash]
./scripts/run_ablations.sh
\end{lstlisting}

which runs all four variants sequentially (approximately 4 hours on 8\,$\times$\,H100 80\,GB), writes their JSON/CSV results, and regenerates the ablation comparison plots. The deterministic hash-based train/val split and the calibrated per-dimension target normalization ensure that every run is independent of random seed for the parts of the pipeline that should be deterministic.

\end{document}